\documentclass{article}

\usepackage[preprint]{neurips_2026_vericode}

\usepackage[utf8]{inputenc} 
\usepackage[T1]{fontenc}    
\usepackage{hyperref}       
\usepackage{url}            
\usepackage{booktabs}       
\usepackage{amsfonts}       
\usepackage{nicefrac}       
\usepackage{microtype}      
\usepackage{xcolor}         

\usepackage{amssymb}
\usepackage{amsmath}
\usepackage{algorithm}
\usepackage{algpseudocode}
\usepackage{tikz}
\usepackage{graphicx}

\makeatletter
\def\ftype@algorithm{4}
\makeatother

\usetikzlibrary{positioning, arrows.meta, fit, calc,
shapes.geometric, backgrounds}

\title{Weave of Formal Thought}

\author{%
  Alexandre~Bouayad \\
  \texttt{alex.bouayad@gmail.com}
}

\newcommand{\githuburl}{https://github.com/alexbouayad/formal}

\begin{document}
\maketitle

\begin{abstract}
  Large language models~(LLMs) attain remarkable surface fluency on
code, yet they do not formally guarantee the syntactic validity of
their output, nor do they typically leverage the hierarchical
structure that defines the target language. While existing
constrained-decoding frameworks (e.g., XGrammar, Guidance, Outlines,
SynCode) offer a solution to the former, they predominantly operate
under rigid assumptions that preclude critical lexical mechanisms
relied upon by modern parsers ---~including context-sensitive lexing
(e.g., Pythonic indentation), maximal-munch tokenization, and keyword
extraction~--- and only approximate the masking of invalid subword
tokens, sacrificing completeness. For the latter, modern code LLMs
inject grammatical structure during training through predetermined
policies rather than learning which structural information to expose.
In this work, we introduce \emph{Weave of Formal Thought}~(WoFT), an
overarching paradigm that unites rigorous syntactic validation with
learned structural representations. First, we present a formal engine
and constrained decoder that is sound and complete with respect to
the full Tree-sitter specification by augmenting generalized LR~(GLR)
parsing with a novel \emph{speculative-lexing} construction that
maintains concurrent lexer-state hypotheses synchronized with the GLR
graph-structured stack; the decoder admits every subword token that
extends to a valid program prefix and rejects every token that does
not. Second, we present a latent-variable fine-tuning method that
trains the language model to interleave non-terminal grammar symbols
directly into the generation process. Utilizing the reweighted
wake-sleep~(RWS) algorithm to optimize the importance-weighted
evidence lower bound~(IW-ELBO) of the surface text, the model learns
to selectively retain formal derivations as an adaptive structural
scratchpad. Across ten diverse programming languages spanning multiple
paradigms (Python, C, C\#, Go, Java, Rust, TypeScript, Haskell, Bash, HTML),
fine-tuning StarCoder2-3B with our RWS objective consistently
outperforms text-only SFT baselines, achieving up to a 14.9\% relative
reduction in surface-token cross-entropy (11.6\% average at convergence,
14.6\% at iso-step), demonstrating that discretionary latent syntax
recovers critical structural information that flat autoregressive
training discards. Our code and implementation are publicly available at
\url{\githuburl}.

\end{abstract}

\section{Introduction}
Modern autoregressive large language models~(LLMs) trained on code attain
remarkable surface fluency: they reproduce idiomatic syntax, identifier
conventions, and library-specific patterns with such fidelity that the generated
text is, on average, statistically indistinguishable from human-written
programs. Yet this competence is brittle in two distinct senses. First, fluency
is not formal correctness: even a state-of-the-art code model produces output
that fails to parse, type-check, or compile at non-trivial rates, necessitating
downstream filtering or repair pipelines whose cost grows with the strictness of
the target specification. Second, and more fundamentally, the model has no
\emph{explicit} access to the hierarchical structure that \emph{defines} the
language. Every program admits a unique, finite, and mathematically rigorous
abstract syntax tree~(AST) derivation; this derivation is precisely the
information that would let the model plan a function before writing its body,
allocate a loop variable before its first use, or close a scope before opening a
new one. Standard autoregressive training, by collapsing the joint distribution
of text and structure onto the marginal over text alone, forces the model to
\emph{re-discover} this hierarchy implicitly at every forward pass.

Recent attempts to inject explicit reasoning into language models have largely
proceeded in the opposite direction. Chain-of-thought~(CoT)
prompting~\cite{chain-of-thought} and its internal variants---Pause
Tokens~\cite{think-before}, Quiet-STaR~\cite{quiet-star}, and continuous-latent
reasoning~\cite{coconut-2025}---grant the model additional intermediate compute,
but the reasoning trajectory itself is either free-form natural language or
unconstrained latent vectors. Neither offers a verifier that can reject
ill-formed reasoning steps, and neither aligns with the discrete, hierarchical
structure that programming languages already provide. The semantic planning gap
is thus a neuro-symbolic gap: the structure of the target language exists,
formally and unambiguously, but is exploited neither at training time nor at
decoding time.

A parallel line of work---constrained decoding---attacks the correctness side of
the problem by masking the vocabulary at each step to the tokens consistent with
a partial parse~\cite{picard, outlines, syncode, xgrammar, xgrammar-2,
llguidance, guidance}. These engines have driven structured generation latency
close to that of unconstrained decoding, but their formal coverage has lagged
behind the languages practitioners actually wish to generate.

To bridge this neuro-symbolic gap, we introduce \emph{Weave of Formal
Thought}~(WoFT), a unified paradigm that integrates rigorous syntactic
validation with learned structural representations. WoFT operates through two
complementary components: a formal inference engine (metaphorically, the
``loom'') and a latent-variable fine-tuning method (the ``weaver'').

First, we present a language-model-agnostic formal engine and constrained
decoder that is sound and complete with respect to the full Tree-sitter
specification~\cite{tree-sitter-2026}. By augmenting Generalized LR~(GLR)
parsing~\cite{tomita-1987} with a speculative-lexing construction, our engine
maintains concurrent lexer-state hypotheses synchronized with a GLR
graph-structured stack. This approach natively supports context-sensitive lexing
(such as Pythonic indentation), maximal-munch tokenization, and declarative
ambiguity resolution. The resulting decoder admits every subword token that
extends to a valid program prefix and rejects every token that does not.

Second, we introduce a latent-variable fine-tuning method that trains the
language model to interleave non-terminal grammar symbols directly into the
generation process. Rather than forcing the model to emit a fixed, deterministic
syntax trace, we treat formal derivations as discrete latent variables. We train
the model using the reweighted wake-sleep~(RWS) algorithm~\cite{rws-2015} to
optimize the importance-weighted evidence lower bound~(IW-ELBO)~\cite{iwae-2016}
of the surface text. This enables the model to utilize formal non-terminals as
an adaptive structural scratchpad, retaining them only when they effectively
compress future surface tokens. Empirically, fine-tuning
StarCoder2-3B~\cite{starcoder2-2024} across ten diverse programming languages
(Python, C, C\#, Go, Java, Rust, TypeScript, Haskell, Bash, HTML) with our RWS
objective consistently outperforms text-only SFT baselines, reducing
surface-token cross-entropy by an average of 11.6\% at final convergence (up to
14.9\% on C) and 14.6\% at iso-step. This demonstrates that discretionary
latent syntax recovers critical structural information that flat
autoregressive training discards. Our complete implementation, including the
formal engine and RWS training pipeline, is publicly available at
\url{\githuburl}.

The remainder of the paper is organized as follows. Section~\ref{sec:related}
reviews related work across constrained decoding, latent structure in language
modeling, and internal reasoning. Section~\ref{sec:formal-engine} details the
design and theoretical guarantees of our formal engine. Section~\ref{sec:woft}
presents our latent-variable formulation, the RWS optimization objective, and
the training architecture. Section~\ref{sec:experiments} details our
experimental setup and reports empirical results on surface token modeling.
Finally, Section~\ref{sec:conclusion} concludes with our research vision and
next steps.

\section{Related Works}
\label{sec:related}
\subsection{Constrained decoding and formal grammars}

Constrained decoding intervenes during autoregressive generation to guarantee
that language model outputs satisfy a formal-language specification, typically
by masking the model's vocabulary at each step to admit only tokens consistent
with a partial parse. PICARD~\cite{picard} introduced incremental parser-guided
decoding for SQL; Outlines~\cite{outlines} and SynCode~\cite{syncode}
generalized this approach to arbitrary regular and context-free languages by
compiling the grammar into finite-state machines and pushdown automata,
respectively, and by precomputing per-token masks offline. Recent engines push
these ideas toward production-scale latency. XGrammar~\cite{xgrammar,
xgrammar-2} partitions the vocabulary into context-independent and
context-dependent subsets to amortize mask construction across decoding steps,
layered on top of byte-level Earley~\cite{earley-1970} and pushdown-automaton
recognizers. Guidance~\cite{guidance} and its Rust backend
LLGuidance~\cite{llguidance} fuse derivative-based parsing directly with the
decoding loop.

A complementary line of work identifies the misalignment between subword
vocabularies and formal language lexemes as a primary source of correctness
failures. DOMINO~\cite{domino-2024} explicitly addresses this gap by combining
offline precomputation with speculative decoding to enforce constraints in a
subword-aligned fashion, achieving near-zero overhead on regex- and CFG-bounded
targets. We share DOMINO's diagnosis but operate one level higher in the
compiler stack: rather than speculating over subword sequences within a fused
regex/CFG state machine, we maintain multiple concurrent lexer states alongside
a Generalized LR~\cite{tomita-1987} graph-structured stack. This preserves a
separate, mature lexer at the granularity of formal lexemes and supports
stateful lexical behaviors that scannerless engines cannot express.

In contrast, our formal engine (Section~\ref{sec:formal-engine}) directly
integrates the full Tree-sitter framework~\cite{tree-sitter-2026}. We
resolve the
subword-alignment gap by maintaining speculative-lexing hypotheses synchronized
with a GLR graph-structured stack, achieving soundness and completeness over
production grammars.

\subsection{Latent structure in language modeling}

The integration of explicit syntactic structure into language modeling has a
rich history in natural language processing. Early structured
language models~\cite{chelba-jelinek,
roark-2001, charniak} demonstrated that conditioning next-token prediction on
partial parse trees improved perplexity. With the advent of deep learning,
frameworks like Recurrent Neural Network Grammars~(RNNGs)~\cite{rnng} and latent
Probabilistic Context-Free Grammars~(PCFGs)~\cite{kim-etal-2019} attempted to
treat these syntactic trees as latent variables. These models optimized the
marginal likelihood of the observed text by marginalizing over all possible tree
structures using dynamic programming (e.g., the Inside-Outside algorithm).

While theoretically elegant, these historical latent-structure models faced
computational scaling limits. Exact marginalization over parse trees scales
cubically with sequence length, making it computationally
intractable for the large vocabularies and context windows of modern
Transformers.
Consequently, the standard language modeling paradigm abandoned explicit latent
syntactic structures, relying entirely on self-attention mechanisms to
implicitly learn flat representations of code and text.

More recent work has pushed internal reasoning further into latent space.
Coconut~\cite{coconut-2025} replaces discrete chain-of-thought tokens with
continuous hidden-state vectors fed back as input embeddings, allowing the model
to encode multiple alternative reasoning steps simultaneously and perform
breadth-first search over reasoning paths. While this circumvents the
textual-coherence overhead of natural-language reasoning, the resulting latents
are continuous vectors with no symbolic semantics; they cannot be validated,
interpreted, or composed by an external verifier. Our framework occupies a
complementary point in the design space: latents are discrete non-terminal
symbols drawn from a formal grammar, and every latent trajectory is verified
against a Tree-sitter GLR oracle. This trades Coconut's continuous geometry for
symbolic groundedness, hierarchical interpretability, and the ability to
mathematically constrain the reasoning trajectory to valid AST derivations.
By modeling formal AST derivations as discrete latents trained via
reweighted wake-sleep~(RWS), our framework (Section~\ref{sec:woft}) enables
autoregressive Transformers to explicitly leverage hierarchical
syntax without the
computational bottleneck of earlier structured models.

\subsection{Internal chain-of-thought and reasoning}

Standard chain-of-thought~(CoT) prompting~\cite{chain-of-thought} improves the
semantic planning of language models by allocating additional
intermediate compute
steps (forward passes) before yielding a final answer. Recently, there has been
a push toward internal or implicit reasoning---allowing models to ``think'' in a
hidden scratchpad that is omitted from the final output. Techniques like Pause
Tokens~\cite{think-before} insert dummy tokens to delay generation, while
Quiet-STaR~\cite{quiet-star} trains models via REINFORCE to generate hidden
natural language rationales, rewarding internal thoughts that minimize the
cross-entropy of future target tokens.

Current internal reasoning frameworks rely heavily on free-form, unconstrained
natural language. Because the hidden rationales are generated autoregressively
without formal boundaries, they remain susceptible to semantic drift,
hallucination, and an unguided search space. When writing code, unconstrained
internal thoughts often fail to align with the strict topological requirements
of the target algorithm.

Our framework redefines internal reasoning by replacing unconstrained natural
language with strict formal grammars. By selectively retaining non-terminal
symbols (e.g., \texttt{<statement>}, \texttt{</statement>}) via our proposal
head, the model engages in latent syntactic planning. These non-terminals act as
a rigorously structured internal scratchpad derived from the true
abstract syntax
tree of the code, grounding the internal reasoning process in valid syntactic
blueprints.

\section{The Formal Engine}
\label{sec:formal-engine}
The first component of the Weave of Formal Thought~(WoFT) paradigm is the formal
engine, which serves as the rigorous structural mechanism---metaphorically, the
``loom''---that guarantees syntactic validity during generation. To successfully
weave formal grammar tokens and surface text without tangling or breaking the
underlying syntactic constraints, the inference engine must precisely track the
grammatical state of the program. In this section, we describe the design of our
formal engine, a language-model-agnostic constrained decoder that achieves
soundness and completeness with respect to the full Tree-sitter specification.

\subsection{GLR parsing and Tree-sitter}

Deterministic LR parsers~\cite{knuth-1965} are limited to unambiguous
context-free grammars. When parsing non-deterministic grammars with shift/reduce
or reduce/reduce conflicts, Generalized LR~(GLR) parsing~\cite{tomita-1987}
bifurcates the parse stack to explore conflicting grammatical hypotheses in
parallel, merging stacks when they reconverge or terminating them when they
prove invalid.

Modern programming languages such as Python, JavaScript, and C++ require parsing
mechanisms beyond pure context-free formalisms. These include lexical context
sensitivity (e.g., disambiguating keywords from identifiers based on parse
state), maximal-munch tokenization, and complex external lexing rules like
indentation tracking. Modern parsers and parser generators integrate these
advanced lexical mechanisms alongside GLR parsing to handle production source
code.

Interactive software environments further require real-time, incremental parsing
to maintain syntax trees during live editing~\cite{incremental-parsing-1997}.
Tree-sitter~\cite{tree-sitter-2026} provides an incremental GLR parsing
infrastructure that rapidly updates syntax trees as code is modified.

However, while Tree-sitter supports incremental parsing for complete lexical
edits in a live editor, language models generate code autoregressively via
arbitrary subword tokens that split lexical boundaries. Because Tree-sitter's
underlying lexer and external scanners rely on complete token streams to enforce
maximal-munch tokenization, they cannot evaluate whether an incomplete subword
token represents a valid prefix of a future legal statement. To serve as an
online constrained decoder, our formal engine extends Tree-sitter's GLR
framework with a speculative, subword-level execution model.

\subsection{Architecture}
\label{sec:engine-arch}

The formal engine is structured as seven co-operating components designed to
reconcile subword-level language model generation with compiler-grade lexical
and syntactic parsing: a \textsc{buffer}, a dual-DFA \textsc{lexer}, an external
\textsc{scanner}, a graph-structured stack (\textsc{stack}), a parse
\textsc{lattice}, an extended GLR \textsc{parser}, and the engine
\textsc{frontend}. The architecture is language-model- and grammar-agnostic:
the \textsc{frontend} decodes emitted subwords into characters and feeds them
to the \textsc{buffer}, while the remaining components operate directly over
compiled Tree-sitter tables and external scanners without modification.

The primary algorithmic hurdle in constrained decoding for production grammars
is the misalignment between subword boundaries and formal lexemes. Traditional
compiler lexers operate under a maximal-munch assumption on complete input
streams, whereas autoregressive decoding feeds arbitrary character fragments.
Our engine resolves this through a \emph{speculative-lexing} mechanism mediated
by the \textsc{lattice} and coroutine-based \emph{scanlets}:

\begin{itemize}
  \item \textbf{Stream buffering and transactional rollbacks}: The
    \textsc{buffer}
    maintains an append-only character stream with transactional checkpointing
    (\texttt{validate} and \texttt{revert}), allowing speculative tokens to be
    evaluated and rejected without corrupting parse state.
  \item \textbf{Dual-track lexing and external scanlets}: The \textsc{lexer}
    advances two parallel DFA states (\texttt{plus} for contextually admissible
      terminals and keywords; \texttt{minus} for keyword-shadowed word
    terminals) to
    reproduce Tree-sitter's keyword extraction. For context-sensitive lexing
    (e.g., Python indentation), the \textsc{scanner} wraps compiled C external
    scanners inside suspendable \emph{scanlets} that yield execution back to the
    parser when hitting the buffer boundary.
  \item \textbf{Lattice of lexico-parse hypotheses}: Rather than maintaining
    only alternative parse stacks via a Graph-Structured
    Stack~(GSS)~\cite{tomita-1987},
    the \textsc{lattice} arranges joint lexico-parse hypotheses in a dependency
    graph. When the buffer is exhausted mid-lexeme, continuation vertices are
    superposed in the lattice, preserving multiple concurrent lexical
    futures and
    enforcing maximal munch across subword boundaries.
  \item \textbf{GLR dispatch and rejection sampling}: The \textsc{parser} drives
    GLR shifts and reductions across active lattice hypotheses via a
    deduplicating
    worklist. At each decoding step, the \textsc{frontend} snapshots
    the lattice,
    speculatively parses candidate subwords, and admits a token if
    and only if at
    least one hypothesis survives.
\end{itemize}
Under standard well-formedness assumptions on the grammar, this construction is
sound and complete with respect to the Tree-sitter specification. We provide the
exhaustive, component-by-component architectural specification and an analysis
of the speculative-lexing mechanics in Appendix~\ref{app:engine-arch}.

\section{Weave of Formal Thought}
\label{sec:woft}
While the formal engine provides the rigorous structural loom to ensure that
language models output valid surface forms at inference time, constrained
decoding alone fundamentally relies on an autoregressive policy that treats
syntax as an external filter rather than an internalized representation. We
hypothesize that modeling the latent syntactic derivation---the \emph{formal
thought} underlying the surface code---provides a richer training signal. To
this end, the second component of the Weave of Formal Thought~(WoFT) paradigm is
a training method that fine-tunes a model to interleave non-terminal grammatical
derivations as discrete latent variables during the generation process. Under
this view, each formal token represents a thought of a \emph{formal} nature, in
the sense that the thought lives in a discretized space---or rather a collapsed
projection of an internal one, continuing the quantum analogy---parameterizing
the structural \emph{form} of the language. We adopt the term \emph{weave} to
reflect that these formal thoughts are not chained sequentially prior to
generation, but are instead interleaved directly into the surface text, acting
as the structural weft that supports the surface code.

\subsection{AST derivations as discrete latents}

To explicitly model the hierarchical structure of the target language, we
construct the abstract syntax tree~(AST) of the surface code using a Tree-sitter
parser and flatten it into a linear sequence that interleaves standard text
subwords with formal grammar tokens. During a depth-first traversal of the AST,
we emit formal tokens that serialize the syntactic derivation: these take the
form of XML-like tags denoting non-terminal grammar rules (e.g.,
\texttt{<statement>}, \texttt{</statement>}) as well as syntactic fields (e.g.,
\texttt{[condition]}, \texttt{[/condition]}). This serialization is
configurable: we can choose to include or exclude field tokens, and to emit only
starting tag tokens (prefix placement), only ending tag tokens (postfix
placement), or both, allowing us to flexibly control the density of the formal
signal injected into the sequence.

The interleaved sequence thus comprises tokens drawn from two logically distinct
vocabularies: the base text vocabulary and the formal grammar vocabulary. We
treat the formal tokens as a sequence of discrete latent variables \( z \),
while the surface text tokens constitute the observed variables \( x \). Our
goal is to maximize the marginal likelihood of the surface code \( p_{\theta}(x)
= \sum_{z} p_{\theta}(x, z) \), effectively marginalizing over all possible
subsets of formal tokens. During training, we represent the decision to keep or
drop each formal token as an independent Bernoulli variable, yielding a sequence
of binary submasks. By selectively masking formal tokens---essentially removing
them from the sequence---we enable the model to explore and marginalize over
different latent derivations.

\subsection{Proposal and generative joint modeling}
\label{sec:modeling}

To optimize the marginal likelihood with discrete latents, we employ an
amortized variational inference framework consisting of two components: a
proposal (or inference) distribution \( q_{\phi}(z \mid x) \) and a generative
model \( p_{\theta}(x, z) \). Rather than maintaining separate networks, both
distributions are parameterized by a single pre-trained base causal language
model augmented with a shared parameter-efficient fine-tuning~(PEFT) adapter.

We extend the vocabulary of the base model by introducing dedicated input and
output embedding layers for the formal tokens. The shared model processes the
mixed sequence of text and formal tokens to produce contextualized hidden
states. From these shared representations, the architecture diverges into two
distinct linear heads.

For the \emph{proposal head}~(\(\phi\)), a one-dimensional linear projection
maps the hidden states of the formal tokens to token-level logits. These logits
parameterize independent Bernoulli distributions over the formal mask, dictating
the probability of keeping each formal token in the sequence. For the
\emph{generative head}~(\(\theta\)), the standard language modeling output
embeddings are used for surface text tokens, while the dedicated formal output
embeddings are used for formal tokens.

By sharing the underlying base model and its PEFT adapter, and routing only
through dedicated linear heads, we efficiently train both \( \phi \) and \(
\theta \) with minimal memory and computational overhead.

\subsection{Optimization via reweighted wake-sleep}
\label{sec:optimization}

While techniques such as the straight-through estimator~(STE)~\cite{ste-2013} or
continuous relaxations, e.g., Gumbel-Softmax~\cite{gumbel-softmax-2017,
concrete-distribution-2017}, are often used to circumvent the
non-differentiability of discrete variables, they are fundamentally incompatible
with our architecture. In our model, the discrete formal mask dictates more than
just the weighting of tokens; it dictates the physical compaction of the
sequence. When a formal token is dropped, all subsequent tokens shift left.
Consequently, this single discrete choice not only alters the overall sequence
length and the positional encodings, but also cascades through the causal
attention mechanism to perturb the logits of every subsequent token. Because
this constitutes discrete stochastic control flow---where the computational
graph itself depends on the sampled
latents~\cite{stochastic-computation-graphs-2016}---continuous relaxations are
inapplicable.

Instead, we optimize the importance weighted evidence lower
bound~(IW-ELBO)~\cite{iwae-2016}. While the standard ELBO bounds the log
marginal likelihood using a single sample, it often forces the proposal
distribution to prematurely collapse around a single mode. The IWAE lower bound
mitigates this by averaging over \( K \) independent particles:
\[
  \mathcal{L}_{\text{IWAE}}
  = \mathbb{E}_{z_1, \dots, z_K \sim q_\phi(z \mid x)}
  \left[ \log \frac{1}{K} \sum_{k=1}^K
    \frac{p_\theta(x_k)}{q_\phi(z_k \mid x)}
  \right]
\]
where \( x_k \) is the sequence with the submask applied. The IWAE objective is
yielding a strictly tighter lower bound that encourages the proposal to explore
multiple plausible trajectories. To optimize this objective, we employ the
reweighted wake-sleep~(RWS) algorithm~\cite{rws-2015}, which provides a highly
effective, low-variance gradient estimator specifically designed for models with
discrete stochastic control flow~\cite{revisiting-rws-2019,decision-iwae-2020}.
We outline one step of the training loop in algorithm~\ref{alg:rws}. Notice that
the IW-ELBO is not differentiated directly: instead, we perform gradient descent
on the surrogate pseudo-loss~\( \mathcal{J}_{\text{RWS}} \). Additionally,
during the wake-\(\theta\) phase, we subsample a single particle proportional to
its normalized importance weight rather than computing the exact weighted sum
over all \( K \) particles. This particle-resampling approach significantly
reduces the memory and computational footprint of the generative backward pass.

\begin{algorithm}[ht]
  \caption{Reweighted Wake-Sleep~(RWS) Training Step}\label{alg:rws}
  \begin{algorithmic}[1]
    \Require Model's proposal parameters \( \phi \) and generative
    parameters \( \theta \)
    \Require Number of particles \( K \)
    \Require Input sequence comprising \( x \) interleaved text and
    formal tokens

    \Statex

    \For{\( k = 1, \dots, K \)}
    \State Sample binary submask \( z_k \sim q_\phi(z \mid x) \)
    \Comment{\textit{Particle Sampling}}

    \State \( x_k \gets \) Apply submask \( z_k \) to \( x \)
    \State \( w_k \gets p_\theta(x_k) /  q_\phi(z_k \mid x) \)
    \EndFor

    \Statex

    \State \( \tilde{w}_1, \dots, \tilde{w}_K \gets \text{Softmax}(\log
    w_1, \dots, \log w_K) \)
    \Comment{\textit{Normalized Weights}}

    \Statex

    \State \( \mathcal{J}_{\text{RWS}} \gets - \sum_{k} \tilde{w}_k
    \log q_\phi(z_k \mid x) \)
    \Comment{\textit{Wake-\(\phi\) Phase}}

    \State Update \( \phi \) via gradient descent on
    \(\mathcal{J}_{\text{RWS}} \)

    \Statex

    \State Subsample single particle \( k' \sim
    \text{Categorical}(\tilde{w}_1, \dots, \tilde{w}_K) \)
    \Comment{\textit{Wake-\(\theta\) Phase}}

    \State \( \mathcal{J}_{\text{RWS}} \gets - \log p_\theta(x_{k'}) \)

    \State Update \( \theta \) via gradient descent on
    \(\mathcal{J}_{\text{RWS}} \)
  \end{algorithmic}
\end{algorithm}

The algorithm relies strictly on the two \emph{wake} phases (wake-\(\theta\) and
wake-\(\phi\)), fully omitting the \emph{sleep} phase present in the original
wake-sleep algorithm. In standard settings, the sleep phase is usually dropped
entirely because training the proposal on data ``dreamed'' by an untrained
generative model introduces a detrimental data distribution
bias~\cite{revisiting-rws-2019}. While our situation is distinct since we are
fine-tuning a pre-trained language model rather than training from scratch, we
still omit the sleep phase because data sparsity is not a primary bottleneck in
our setting. The model converges rapidly relative to the abundance of available
surface code, rendering the sleep phase unnecessary for basic convergence.
Nonetheless, future iterations could incorporate the sleep phase as a
regularizer, driving the proposal to better approximate the generative
posterior; this may lower the variance of the importance weights in the wake
phase, thereby mitigating the self-normalized importance sampling~(SNIS) bias
inherent to the wake-\(\phi\) update.

Compared to alternative discrete score-function estimators such as
VIMCO~\cite{vimco-2016}, ARM~\cite{arm-2019}, DISARM~\cite{disarm-2020}, and
OVIS~\cite{ovis-2020}, which scale gradient updates by unnormalized objective
magnitudes and risk signal collapse near sharp local optima, RWS maintains a
constant total gradient weight of~$1$. This provides a natural trust-region
regularization that yields low-variance, stable updates under
discrete stochastic
control flow. We provide an extended theoretical comparison of these
estimators in
Appendix~\ref{app:estimators}.

\section{WoFT Improves Surface Modeling}
\label{sec:experiments}
In this section, we empirically evaluate whether Weave-of-Formal-Thought
fine-tuning via RWS improves the base language model's ability to model surface
tokens across diverse programming languages. We first detail the experimental
setup, including the multi-language corpus, tokenization strategy, and
optimization hyperparameters. We then present empirical results comparing the
generative performance of our approach against standard supervised fine-tuning
baselines on Python, followed by a cross-lingual evaluation across nine
additional languages.

\subsection{Experimental setup}

We conduct our experiments using StarCoder2-3B~\cite{starcoder2-2024}
as the base
architecture. To evaluate generality across programming paradigms, we train
models across ten distinct languages: Python, C, C\#, Go, Java, Rust,
TypeScript, Haskell, Bash, and HTML. For each language, our training dataset consists
of a 15,000-sequence subset sampled from The Stack~v2~\cite{starcoder2-2024},
specifically targeting source files between 1 and 100,000 bytes. To
process these
files into formal sequences, we rely on StarCoder2-3B's subword
tokenizer for the
underlying text. We parse raw source code using Tree-sitter to construct an
abstract syntax tree~(AST), then traverse this AST in prefix order, interleaving
surface text tokens with newly introduced formal tokens denoting non-terminal
grammar rules. This extended formal vocabulary is appended directly to the base
tokenizer. Finally, the sequences are processed with a maximum length of 16,000
tokens.

To efficiently accommodate the newly introduced formal grammar vocabulary and
enable the model to internalize syntactic structure, we employ low-rank
adaptation~(LoRA)~\cite{lora-2022} with rank \( r = 32 \). Adapters are
applied across all attention projections (\texttt{q\_proj}, \texttt{k\_proj},
\texttt{v\_proj}, \texttt{o\_proj}), feed-forward layers (\texttt{c\_fc},
\texttt{c\_proj}), and crucially, the embedding and language modeling heads
(\texttt{embed\_tokens}, \texttt{lm\_head}).

All models are optimized using AdamW~\cite{adamw-2019} with a learning rate of
\( 3 \times 10^{-4} \), weight decay of \( 0.01 \), and gradient clipping at
maximum norm \( 1.0 \). Training is performed on an NVIDIA A40 GPU in
\texttt{bfloat16} precision, leveraging PyTorch's~\cite{pytorch-2019} Scaled
Dot-Product Attention~(SDPA)~\cite{sdpa-2022, sdpa-2-2024} and gradient
checkpointing~\cite{gradient-checkpointing-2016}. For WoFT models trained via
reweighted wake-sleep, we use \( K = 4 \) independent particles per sequence to
construct the variance-reduced gradient estimators. Our implementation is built
upon the Hugging Face \texttt{transformers}~\cite{hf-transformers-2020},
\texttt{datasets}~\cite{hf-datasets-2021}, and \texttt{peft}~\cite{hf-peft-2023}
libraries. The complete codebase is publicly available at \url{\githuburl}.

\subsection{Generative performance across programming languages}

To evaluate the performance of Weave-of-Formal-Thought~(WoFT) fine-tuning, and
to assess the efficacy of our training method, we compare the generative
performance of our model trained via the reweighted wake-sleep~(RWS) algorithm
against standard supervised fine-tuning~(SFT) baselines. All models are trained
to optimize next-token prediction, measured via per-token cross-entropy loss.

\paragraph{Isolating latent structure on Python.}
To isolate the effect of our discrete latent variable formulation from the mere
presence of syntactic data, we first evaluate two baselines on Python: a
\emph{Text SFT} baseline trained exclusively on the original surface text, and
a \emph{Text+Formal SFT} baseline trained on surface text interleaved with
formal grammar tokens via standard autoregressive teacher-forcing. Following
standard single-epoch training on deduplicated data, the smoothed online
training loss (250-step running average) provides an unbiased estimator of
generalization error.

\begin{figure}[ht]
  \centering
  \includegraphics[width=0.64\linewidth]{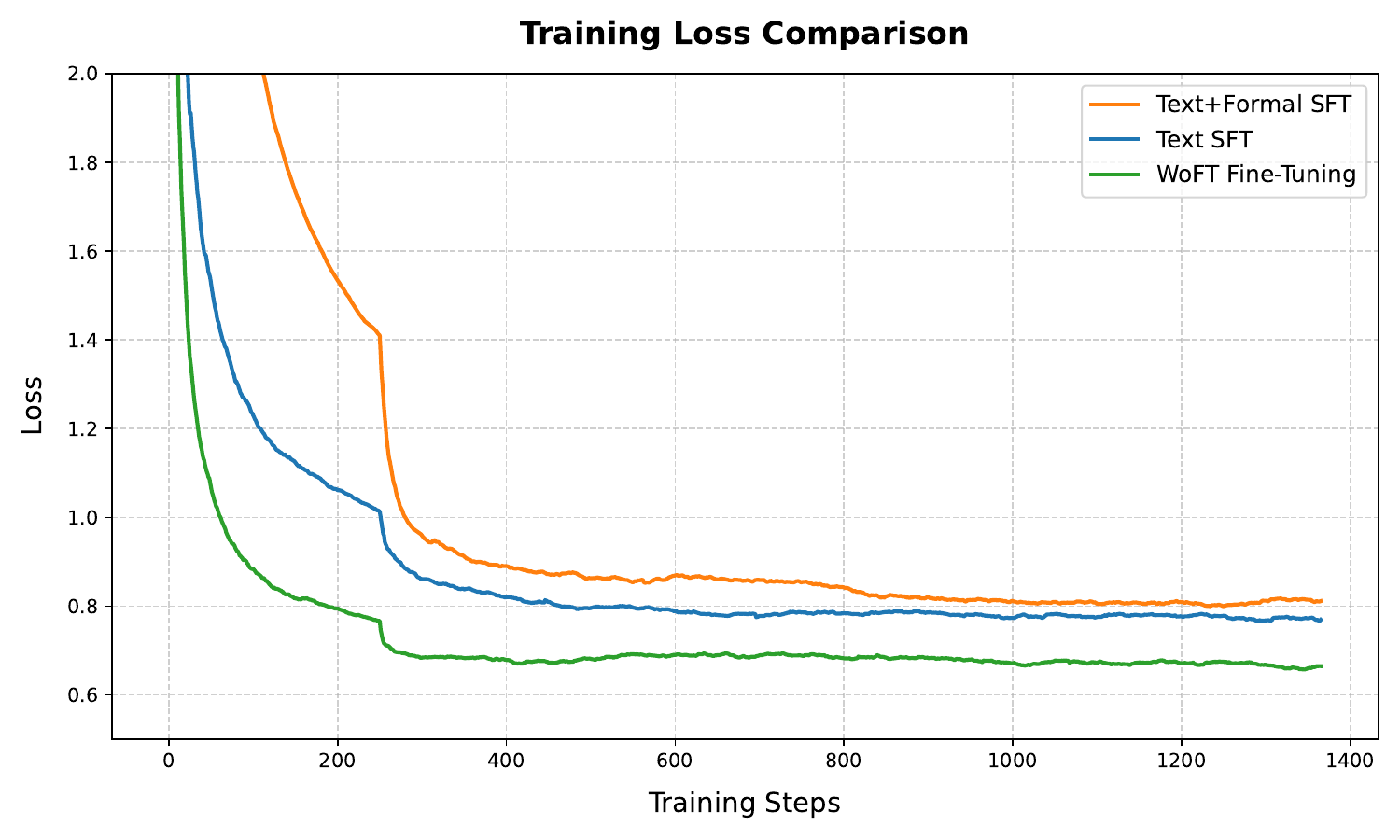}
  \caption{Smoothed online training loss (per-text-token cross-entropy) across
    the three fine-tuning methods on Python (running average over 250 steps).
    WoFT fine-tuning via RWS achieves a substantially lower surface
    cross-entropy
  (0.66) than both Text SFT (0.77) and Text+Formal SFT (0.82).}
  \label{fig:wandb}
\end{figure}

As shown in Figure~\ref{fig:wandb}, training WoFT with the RWS objective yields
a substantial reduction in surface cross-entropy on Python, converging to
\(0.66\) compared to \(0.77\) for Text SFT (a 13.6\% relative reduction).
Crucially, Text+Formal SFT performs worse (\(0.81\)), confirming that forcing
the model to deterministically unroll trivial syntax via teacher forcing
distracts capacity from high-entropy surface tokens. In contrast, WoFT
selectively retains formal derivations only when they effectively
compress future
surface text, serving as an adaptive structural scratchpad.

\paragraph{Cross-lingual validation across ten languages.}
To verify that these structural benefits generalize beyond Python, we fine-tune
StarCoder2-3B using WoFT and Text SFT across nine additional languages
spanning diverse programming paradigms: C (imperative/systems), C\#
(object-oriented), Go (concurrent/systems), Java (object-oriented), Rust
(functional/systems), TypeScript (web/multi-paradigm), Haskell (purely
functional), Bash (shell scripting), and HTML (declarative/markup).

\begin{table}[ht]
  \caption{Generative cross-entropy of StarCoder2-3B fine-tuned via
    Text SFT vs.\
    Weave of Formal Thought (WoFT) across ten programming languages
    from The Stack~v2. WoFT consistently outperforms Text SFT across every
    language, achieving an average relative loss reduction of 11.6\% at final
    convergence and 14.6\% at iso-step.}
  \label{tab:multilang}
  \centering
  \small
  \begin{tabular}{llcccc}
    \toprule
    Language & Paradigm & Text SFT & WoFT (RWS) & $\Delta$ Final (\%)
    & $\Delta$ Iso-step (\%) \\
    \midrule
    Python     & Multi-paradigm / Scripting & 0.7691 & 0.6648 &
    13.57\% & 13.57\% \\
    C          & Imperative / Systems       & 0.6902 & 0.5874 &
    14.90\% & 15.95\% \\
    C\#        & Object-Oriented            & 0.5782 & 0.5205 &
    \phantom{0}9.98\% & 15.77\% \\
    Go         & Concurrent / Systems       & 0.6021 & 0.5273 &
    12.43\% & 15.70\% \\
    Java       & Object-Oriented            & 0.5856 & 0.5204 &
    11.14\% & 14.81\% \\
    Rust       & Functional / Systems       & 0.5526 & 0.4929 &
    10.80\% & 15.01\% \\
    TypeScript & Multi-paradigm / Web       & 0.6526 & 0.5911 &
    \phantom{0}9.43\% & 12.54\% \\
    Haskell    & Purely Functional          & 0.8997 & 0.8195 &
    \phantom{0}8.92\% & 15.08\% \\
    Bash       & Shell Scripting            & 1.1311 & 0.9940 &
    12.12\% & 12.57\% \\
    HTML       & Declarative / Markup       & 0.9414 & 0.8206 &
    12.83\% & 14.97\% \\
    \midrule
    \textbf{Average} & ---                  & \textbf{0.7403} &
    \textbf{0.6538} & \textbf{11.61\%} & \textbf{14.60\%} \\
    \bottomrule
  \end{tabular}
\end{table}

Table~\ref{tab:multilang} reports the results. WoFT achieves strictly lower
cross-entropy across every single language without exception. At final
convergence, relative reductions average 11.6\%, ranging from 8.9\% on Haskell
to 14.9\% on C. Notably, each gradient step corresponds to the exact same
number of sequences trained on across both methods (10 sequences per step via
gradient accumulation). What emerges is that WoFT converges on average three times
faster. Evaluated at iso-step (comparing Text SFT at the exact same
number of sequences trained on as WoFT's final step), WoFT's relative loss
reduction averages 14.6\% across all languages (spanning 12.5\% to 16.0\%).
This establishes that the benefits of learned
discretionary latent syntax are paradigm-invariant and universally applicable,
providing superior sample efficiency alongside stronger generative modeling.

\section{Conclusion}
\label{sec:conclusion}
In this work, we introduced two complementary contributions to bridge the
neuro-symbolic gap in code generation. First, we developed a formal inference
engine that provides sound and complete constrained decoding with respect to the
full Tree-sitter specification, resolving the subword-lexeme
misalignment through
speculative GLR parsing. Second, we proposed Weave of Formal Thought~(WoFT), a
latent-variable fine-tuning method that trains language models to interleave
non-terminal grammar derivations as discrete latents optimized via the
reweighted wake-sleep~(RWS) algorithm.

Our empirical findings
across ten diverse programming languages demonstrate that treating formal
syntax as an adaptive structural scratchpad consistently reduces surface-token
cross-entropy (averaging 11.6\% at final convergence and 14.6\% at iso-step) on
StarCoder2-3B compared to text-only SFT baselines, while teacher-forcing full
deterministic syntax traces actively degrades surface modeling.

In the near term, practical next steps focus on low-level decoder acceleration
(byte-level lexing, trie-based masking, incremental parsing), refined empirical
ablations, functional execution benchmarks, and online coupling of the formal
engine with WoFT fine-tuning.  We detail this roadmap and our broader
research vision
in Appendix~\ref{app:future-work}.

Ultimately, our results confirm that explicitly modeling syntax assists surface
generation when the model is granted the discretion to prune unneeded
structure. This aligns closely with
established findings in cognitive science and neuroscience, where experimental
evidence demonstrates that the human brain constructs hierarchical,
grammatical hypotheses during real-time language processing. By
treating formal syntax as both an external guarantee and
an internal latent reasoning substrate, this work establishes a grounded path
toward bridging structural reasoning and statistical fluency.

\begin{ack}
  The author thanks G-Research for their support and authorization to
publish this work. The author is also deeply grateful to Inria Paris
for their welcoming hospitality and supportive research environment.

\end{ack}

\bibliographystyle{unsrt}
\bibliography{references}

\appendix
\section{Detailed Formal Engine Architecture}
\label{app:engine-arch}

In this section, we provide the full architectural specification of the seven
co-operating components comprising the formal engine introduced in
Section~\ref{sec:formal-engine}: the stream buffer, the DFA-based lexer, the
stream scanner, the graph-structured stack, the parse lattice, the extended GLR
parser, and the engine frontend.

\paragraph{\textsc{buffer}.} At the upstream end of the engine, the
\textsc{buffer} exposes a single-character lookahead reader over an append-only
stream of characters. The \texttt{feed} operation appends a chunk of text, or an
end-of-stream~(EOS) sentinel, without disturbing the current read offset; this
is how the engine ingests the characters of a subword token. The
\texttt{validate} and \texttt{revert} operations together implement a checkpoint
mechanism: once a token has been accepted by the \textsc{parser}, the
\textsc{buffer} is validated and its checkpoint advanced to the current write
head; if the token is rejected, the \textsc{buffer} is reverted and all
characters written past the last checkpoint are discarded. The \textsc{buffer}
owns the text; both the \textsc{lexer} and the \textsc{scanner} operate on it
through a small reader interface exposing offset, lookahead, and the
at-end-of-file predicate.

\paragraph{\textsc{lexer}.} The \textsc{lexer} is a deterministic finite-state
recognizer driven by a single compiled transition table. It advances two DFA
states in parallel, \texttt{plus} and \texttt{minus}, which share the same
transition function but are seeded from distinct starting states that depend on
the parsing context. The \texttt{plus} state recognizes the terminals that are
valid lookaheads in the current parse state, including any keywords and the
generic word terminal whenever the latter is itself admissible. The
\texttt{minus} state is seeded when the word terminal is \emph{not} a valid
lookahead, and while at least one keyword is a valid lookahead, in which case it
recognizes the word terminal alone. Together these two states give the engine
the information it needs to reproduce Tree-sitter's keyword extraction. At every
step the \textsc{lexer} reads one character, advances the two DFA states, and
records the most recent accepting position as a candidate endpoint, yielding
maximal-munch tokenization. The \textsc{lexer}'s \texttt{lex} operation drives
this loop from the current reader position: it steps both DFA tracks until both
die or the \emph{buffer} is exhausted, returns the last matched symbol together
with the accepting position, and writes the surviving DFA states back into the
\textsc{lexer}'s state before returning. When both DFA states accept at a shared
endpoint, \texttt{plus} acceptance takes precedence over \texttt{minus}
acceptance. The pair of DFA states constitutes the \textsc{lexer}'s state, which
the \textsc{parser} can read and reset to suspend and resume lexing at subword
token boundaries when the \textsc{buffer} is exhausted mid-lexeme.

\paragraph{\textsc{scanner}.} Tree-sitter grammars routinely rely on external
scanners written in~C to recognize tokens that cannot be expressed as regular
languages, e.g., indentation in Python, heredocs, raw strings, template
literals. The engine integrates such scanners directly through a foreign
function interface: it loads the compiled scanner library and exposes the
Tree-sitter callback surface. Across scans, the \textsc{scanner}'s state is
persisted through Tree-sitter's native serialization protocol: at the end of
each scan the state is serialized into a byte string, which the engine carries
alongside the \textsc{lexer}'s state and restores at the start of the next scan.
Within a single scan, however, a non-trivial challenge arises because external
scanners are written under the assumption of a complete input: when the
\textsc{scanner} requests the next character past the end of the currently
available text, it cannot simply block. We resolve this by running each external
scanner invocation inside a \emph{scanlet}, a suspendable coroutine carrying one
external scanner call, which yields control back to the \textsc{parser} when the
the external scanner hits the end of the \textsc{buffer}. The \textsc{scanner}'s
\texttt{scan} operation either resumes an existing scanlet or spawns a fresh one
at the current \textsc{buffer} position, seeded from the serialized state and
the set of recognizable external symbols, then switches execution into the
scanlet; the scanlet runs the C scanner callback directly, yielding back to the
\textsc{parser} whenever the callback requests a character past the
\textsc{buffer} boundary, and returns the matched symbol together with the
accepting position once the callback completes. Once additional text has been
fed, the same scanlet is resumed and continues from exactly the C-side
instruction at which it was suspended. Cloning a scanlet duplicates its
suspended state, allowing each branching parse hypothesis to carry its own
in-flight scan.

\paragraph{\textsc{stack}.} The parser operates over a graph-structured
stack~(GSS), the standard device for sharing structure across the parallel
stacks of a GLR parser. The GSS is a directed acyclic graph, where each path
corresponds to one stack. Each node of the GSS carries a parse state and a set
of predecessor links. The \texttt{push} operation takes a parse state and embeds
it into a new node that points back to the top of one of the parallel stacks.
Two nodes carrying the same parse state can be merged by taking the union of
their predecessor sets, so parse paths that converge on a common parse state can
be represented by a single shared node, while the distinct histories stay
separate. The \texttt{pop} operation, parameterized by a count, returns the set
of nodes reachable by traversing backwards that many links, so the reduction of
a grammar rule can fan out over multiple paths, without ever materializing the
underlying parse forest.

\paragraph{\textsc{lattice}.} A graph-structured stack captures alternative
parse stacks, but the engine must also account for alternative lexical
futures---suspended \textsc{lexer} and \textsc{scanner} states that may resolve
to a lexical token. The engine must additionally track the ordering constraints
that maximal-munch tokenization imposes on such futures, and it must keep each
lexical hypothesis synchronized with the parse stack it belongs to. This is
realized by the \textsc{lattice}, which consists of a superposition of
lexico-parse hypotheses arranged as a dependency graph. Each lattice vertex
records an alternative parse stack, a \textsc{buffer} position, a \textsc{lexer}
state, and a \textsc{scanner} state. The \textsc{scanner} state is either a
serialized Tree-sitter state or a live scanlet when an external scan is in
flight. A vertex is keyed by a \emph{signature} consisting of its top parse
state, text position, \textsc{lexer} state, and \textsc{scanner} state. Vertices
sharing a parent vertex are grouped by their signatures and merged by taking the
union of their parse stacks, so lexico-parse branches that reconverge never
multiply. The \textsc{lattice} is accessed through a set of \emph{heads} that
point to its vertices. The \texttt{focus} operation designates as
\emph{visible}---the active target of lattice operations---the vertex pointed to
by a head. The \emph{superpose} operation inserts a new vertex between the
visible vertex and its parent; its destructive dual, \texttt{collapse}, tears
down the subtree beneath the visible vertex. The \texttt{create} operation
spawns a new vertex under the same parent as the visible vertex; its destructive
dual, \texttt{annihilate}, removes the visible vertex---reconnecting its
children to its parent. The two constructive operations return a fresh head
pointing to the new vertex, while the two destructive operations retire the head
pointing to the vanished vertex.

\paragraph{\textsc{parser}.} The \textsc{parser} is a GLR parser over the
Tree-sitter parse and goto tables, where the collection of parse
hypotheses---pairs of a stack version and a buffer position---is extended into
the richer DAG structure of the \textsc{lattice}. Every \texttt{parse} step
opens with a lex step that yields a single potential lookahead by dispatching,
in order, to the \textsc{scanner}'s \texttt{scan} or the \textsc{lexer}'s
\texttt{lex} operation. If an external scan is already in flight, the scanlet is
resumed; if the \textsc{lexer} is in a suspended state, it is similarly resumed;
otherwise the \textsc{scanner} is tried first, and, if it yields nothing, the
\textsc{lexer} is tried---the seed for both is taken from the \emph{lex mode} of
the current parse state. When the \textsc{buffer} is exhausted mid-lexeme---that
is, when the \textsc{scanner} returns a scanlet or when the \texttt{plus} DFA
track of the \textsc{lexer} is still alive---the parser inserts a continuation
vertex into the \textsc{lattice} via the \texttt{superpose} operation, carrying
the in-progress scanlet or the \textsc{lexer}'s state so that scanning or lexing
can be resumed verbatim once more text arrives. A lexed lookahead (or the
\textsc{lexer} ending in a \texttt{minus}-only state; see below) triggers the
\texttt{collapse} of the visible parse vertex. The parse vertex is eventually
and unconditionally \texttt{annihilate}d, and if a lookahead is present, the
\textsc{parser} then drives shifts and reductions. Reductions are serialized via
a deduplicating worklist, fanning out across the graph-structured stack at
parse-table conflicts and merging derivations that reconverge on a common stack
top; a shift \texttt{create}s a new parse vertex. The \textsc{parser} cycles
through the \textsc{lattice} until all hypotheses are either exhausted or paused
at the \textsc{buffer} boundary. The \textsc{lattice} is therefore the locus of
the \emph{speculative-lexing} mechanism: within a single parse cycle the parser
may commit to a token at one endpoint while a parent hypothesis continues lexing
past it, preserving the maximal munch property even when a subword token
boundary falls beyond the characters currently in the \textsc{buffer}.

When the \textsc{lexer} returns the word symbol from \texttt{minus} acceptance,
the terminal is inadmissible in the current parse state; the parse node is
therefore \texttt{collapse}d and \texttt{annihilate}d and no parse action is
triggered, integrating Tree-sitter's keyword extraction---an identifier that
merely begins with a keyword is lexed to its full extent and rejected, while a
lexeme equal to a contextually valid keyword is shifted as that keyword. When
the \textsc{lexer} ends in a \texttt{minus}-only state at the \textsc{buffer}
boundary, the \textsc{parser} conservatively retires the hypothesis via
\texttt{collapse} and \texttt{annihilate}. This is the only over-approximation
the engine makes to the filtering of subword tokens; the filter remains
complete for most programming languages, essentially complete for the others,
and could be lifted entirely.

\paragraph{\textsc{frontend}.} The \textsc{frontend} integrates the six
components above into the standard LLM generation loop via rejection sampling.
Before each subword token is emitted, the \textsc{frontend} snapshots the
current \textsc{lattice}. The language model produces next-token logits; the
\textsc{frontend} takes the highest-scoring candidate token, decodes it into
characters, feeds them to the \textsc{buffer}, and runs the \textsc{parser}. If
at least one hypothesis of the \textsc{lattice} survives, the token is
accepted: the \textsc{buffer} is \texttt{validate}d and the \textsc{frontend}
advances. If every hypothesis dies, the token is rejected: the \textsc{buffer}
is \texttt{revert}ed, the \textsc{lattice} is restored from its snapshot,
the offending token's logit is set to~\(-\infty\), and the \textsc{frontend}
resamples from the updated distribution.

\section{Comparison with Alternative Discrete Latent Estimators}
\label{app:estimators}

To appreciate our choice of the reweighted wake-sleep~(RWS) algorithm in
Section~\ref{sec:optimization}, it is instructive to compare it against
alternative score-function estimators developed for discrete latent variables.

Unbiased estimators such as VIMCO~\cite{vimco-2016} reduce variance via
leave-one-out baselines:
\[
  \hat{\nabla}_{\phi}^{\text{VIMCO}}
  = \sum_{k=1}^K \left[ \log \hat{p}(x) - \log \hat{p}_{-k}(x) \right]
  \nabla_{\phi} \log q_{\phi}(z_k \mid x)
\]
where \( \hat{p}(x) = \frac{1}{K}\sum_{j} w_j \) and \( \hat{p}_{-k}(x) \) is a
leave-one-out marginal likelihood estimate. While unbiased, VIMCO suffers from
decaying signal-to-noise ratios as the number of particles \( K \)
increases~\cite{rainforth-2018}, limiting its utility in multi-sample settings.

More recently, estimators such as ARM~\cite{arm-2019} and
DISARM~\cite{disarm-2020}
leverage antithetic sampling over the uniform variables underlying
the Gumbel-Max
trick to achieve strictly lower variance for categorical latents. In parallel,
OVIS~\cite{ovis-2020} derives the theoretically optimal control variate by
minimizing the variance of the gradient estimator directly. However, because
these estimators scale their gradient updates by quantities that correlate with
the magnitude of the unnormalized objective function, they frequently
suffer from
signal collapse and severe gradient variance near sharp local optima.

In contrast, the wake-\(\phi\) phase of the reweighted wake-sleep algorithm
updates the proposal parameters~\( \phi \) via the surrogate objective:
\[
  \mathcal{J}_{\text{wake-}\phi}
  = - \sum_{k=1}^K \tilde{w}_k \log q_{\phi}(z_k \mid x),
  \quad \text{where} \quad
  \tilde{w}_k = \frac{w_k}{\sum_{j=1}^K w_j}
\]
Because the normalized importance weights satisfy \( \sum_{k=1}^K
\tilde{w}_k = 1 \)
by construction, the total gradient weight assigned to the proposal update is
strictly bounded and invariant to the absolute magnitude of the underlying
likelihoods. This effectively acts as a natural trust-region regularization of
OVIS, preventing gradient explosion or collapse and ensuring robust convergence
even when discrete choices induce major topological shifts in causal attention.

\section{Next Steps and Research Vision}
\label{app:future-work}

In this section, we detail practical next steps and long-term research horizons.

\subsection{Practical Next Steps}

To further minimize constrained decoding overhead, we plan to integrate advanced
low-level execution mechanisms, including byte-level lexing, trie-based
vocabulary masking, and optimized rejection sampling architectures inspired by
high-performance generation engines like Guidance~\cite{guidance} and
XGrammar~\cite{xgrammar-2}. We also aim to advance lazy compilation by
incorporating derivative-based parsing for lexing and fusing Earley
parsers~\cite{earley-1970} with our GLR lattice parser. In parallel, we intend
to supplement our engine with the highly efficient incremental parsing
capacities offered by Tree-sitter, which parsing based on the GLR algorithm can
natively support~\cite{incremental-parsing-1997}. Finally, we plan to develop a
low-level external stream scanner API to seamlessly accommodate languages
requiring arbitrary external state tracking directly over token streams,
extending Tree-sitter's external scanner API to autoregressive generation.

Future empirical work will conduct extensive ablation studies to isolate the
mechanisms driving our observed loss reductions. Specifically, we will
investigate regimes where full deterministic syntax injection brings no
improvement (confirming that teacher-forcing the complete syntax trace provides
the same or worse modeling capacity than baseline SFT), as well as evaluating
random masking and caching strategies (with and without fixed frequencies) to
verify that the gains stem specifically from learned, adaptive latent routing.
Crucially, we aim to evaluate our fine-tuned models across standard, widely
established benchmarks for code generation to rigorously assess downstream
functional correctness and execution accuracy.

A central near-term objective is the bidirectional coupling of our two
contributions: deploying the formal engine during generation to actively verify
and constrain the model's latent non-terminal proposals. By validating candidate
syntactic non-terminals against the partial GLR parse state before they are
emitted, the engine can prune invalid structural derivation paths, ensuring that
internal syntactic scratchpads remain provably well-formed.

\subsection{Long-Term Research Vision}

\paragraph{Toward AI systems that formally plan.}
Standard autoregressive language models treat highly structured domains---such
as source code, formal logic, and mathematical proofs---as flat,
one-dimensional sequences of tokens. This absence of explicit hierarchical
representations induces well-documented failure modes: syntax hallucinations,
inefficient long-horizon planning, and brittle compositional reasoning. Existing
neuro-symbolic methods typically intervene only at inference time via rigid
rejection sampling, treating formal grammar rules as external constraints rather
than internalizing them during training. We view WoFT as a stepping stone toward
transforming generative language models from surface-level statistical token
predictors into verifiable reasoning systems that natively plan within formal
structural abstractions.

\paragraph{Hierarchical and non-linear generation.}
Autoregressive decoding is traditionally restricted to left-to-right, linear
token generation. However, because formal non-terminal symbols encapsulate
complete, well-formed syntactic subtrees, they provide a natural foundation for
non-linear, hierarchical generation. Rather than predicting every subword token
strictly sequentially, an architecture can plan high-level syntactic scaffolding
(such as function signatures, loop structures, or control branches) before
recursively populating constituent subtrees. Beyond preserving long-range
structural coherence across extended contexts, such hierarchical generation
opens possibilities for parallelized subtree decoding and modular state reuse
across independent syntactic scopes.

\paragraph{Structural working memory and context compression.}
The discrete hierarchical representations learned by WoFT offer a dual
application for memory management in long-horizon reasoning and agentic
workflows. As models execute multi-step software engineering or reasoning
tasks, context windows quickly become saturated with verbose code and execution
traces. Rather than relying on lossy natural-language summaries or ungrounded
latent vectors, models can learn to compress past working memory using the exact
same formal grammar non-terminals. By abstracting completed functional blocks or
verified execution traces into compact abstract syntax tree nodes, the system
retains precise structural and semantic fidelity while drastically reducing
effective context length.

\paragraph{Generalization to dependent type systems.}
While our current formal engine and latent formulation operate over
compiler-grade Tree-sitter grammars, formal reasoning in pure mathematics and
safety-critical software verification requires expressivity beyond context-free
formalisms. A natural frontier is extending the WoFT framework to dependent type
theories, on which interactive theorem provers such as
Lean~4~\cite{lean4-2021} are built.
In formal mathematics, proof construction proceeds through tactic sequences
guided by evolving proof states. By formulating proof
derivations as latent structural variables, this paradigm can help
eliminate deductive
hallucinations and help structure proof search.

\paragraph{Learning latent formal structures.}
In our current framework, formal non-terminals are defined a priori by an
externally supplied grammar specification. A compelling theoretical question is
whether these formal structures can themselves be inferred and learned. By
optimizing the marginal likelihood jointly over grammar production choices and
token sequences, future systems could dynamically induce, adapt, and refine
latent formal grammars for novel Domain-Specific Languages (DSLs) or
under-specified formal domains. This
would for example allow an autonomous agent to synthesize its own
verifiable, discrete
internal languages tailored to specific problem-solving domains.

\paragraph{Continuous and differentiable structural latents.}
Finally, an intriguing frontier involves relaxing discrete grammar symbols into
continuous formal representations while maintaining the
symbolic invariants of the formal engine at structural boundaries. By combining
the fluid, end-to-end differentiability of continuous latent reasoning with the
rigorous boundaries provided by formal grammars, this hybrid approach could
realize fully differentiable yet symbolically grounded syntactic planning.

\end{document}